%% file: main.tex
\documentclass[10pt,twocolumn,letterpaper]{article}

\usepackage[pagenumbers]{cvpr} 

\usepackage{graphicx}
\usepackage{amsmath}
\usepackage{amssymb}
\usepackage{multirow}
\usepackage{booktabs}
\usepackage{ulem}
\usepackage{makecell}

\usepackage[pagebackref,breaklinks,colorlinks,bookmarks=false]{hyperref}

\usepackage[capitalize]{cleveref}
\crefname{section}{Sec.}{Secs.}
\Crefname{section}{Section}{Sections}
\Crefname{table}{Table}{Tables}
\crefname{table}{Tab.}{Tabs.}

\def\cvprPaperID{7174} 
\def\confName{CVPR}
\def\confYear{2022}

\begin{document}

\newcommand{\zizhang}[1]{\textcolor{red}{[Zizhang:#1]}}
\newcommand{\mengmeng}[1]{\textcolor{blue}{[Mengmeng:#1]}}

\def\ie{\textit{i.e.~}}

\newcommand{\netname}{MaIL}

\title{MaIL: A Unified Mask-Image-Language Trimodal Network \\
for Referring Image Segmentation}

\author{
Zizhang Li\footnotemark[1]
\quad
Mengmeng Wang\footnotemark[1]
\quad
Jianbiao Mei
\quad
Yong Liu\footnotemark[2]\\
\small College of Control Science and Engineering, Zhejiang University\\
{\tt\small \{zzli, mengmengwang, jianbiaomei\}@zju.edu.cn,~yongliu@iipc.zju.edu.cn}
}
\maketitle
\renewcommand{\thefootnote}{\fnsymbol{footnote}}
\footnotetext[1]{Equal contribution}
\footnotetext[2]{Corresponding Author}

\input{contents/1-abstract}

\input{contents/2-introduction}
\input{contents/3-related_work}
\input{contents/4-methods_new}
\input{contents/5-experiments}
\input{contents/6-conclusion}

{\small
\normalem
\bibliographystyle{ieee_fullname}
\bibliography{contents/reference}
}



\end{document}

%% file: contents/1-abstract.tex
\begin{abstract}
\label{abstract}
Referring image segmentation is a typical multi-modal task, which aims at generating a binary mask for referent described in given language expressions. Prior arts adopt a bimodal solution, taking images and languages as two modalities within an encoder-fusion-decoder pipeline. 
However, this pipeline is sub-optimal for the target task for two reasons. First, they only fuse high-level features produced by uni-modal encoders separately, which hinders sufficient cross-modal learning. Second, the uni-modal encoders are pre-trained independently, which brings inconsistency between pre-trained uni-modal tasks and the target multi-modal task.
Besides, this pipeline often ignores or makes little use of intuitively beneficial instance-level features.
To relieve these problems, we propose \textbf{\netname}, which is a more concise encoder-decoder pipeline with a \textbf{Ma}sk-\textbf{I}mage-\textbf{L}anguage trimodal encoder. Specifically, \netname\ unifies uni-modal feature extractors and their fusion model into a deep modality interaction encoder, facilitating sufficient feature interaction across different modalities. Meanwhile, \netname\ directly avoids the second limitation since no uni-modal encoders are needed anymore. 
Moreover, for the first time, we propose to introduce instance masks as an additional modality, which explicitly intensifies instance-level features and promotes finer segmentation results. 
The proposed \netname\ set a new state-of-the-art on all frequently-used referring image segmentation datasets, including RefCOCO, RefCOCO+, and G-Ref, with significant gains, 3\%-10\% against previous best methods. Code will be released soon.
\end{abstract}

%% file: contents/2-introduction.tex
\section{Introduction}
\label{intro}
 This paper focuses on referring image segmentation, which is a typical multi-modal task by generating segmentation on images with the guidance of language descriptions. Since the real world contains not only vision but also multiple modalities, like audios and languages, this joint vision-text task could push a step further to better vision-language understanding, providing more user-friendly and convenient functions for practical applications like image manipulation and editing. 

\begin{figure}[t]
  \centering
  \includegraphics[width=\linewidth]{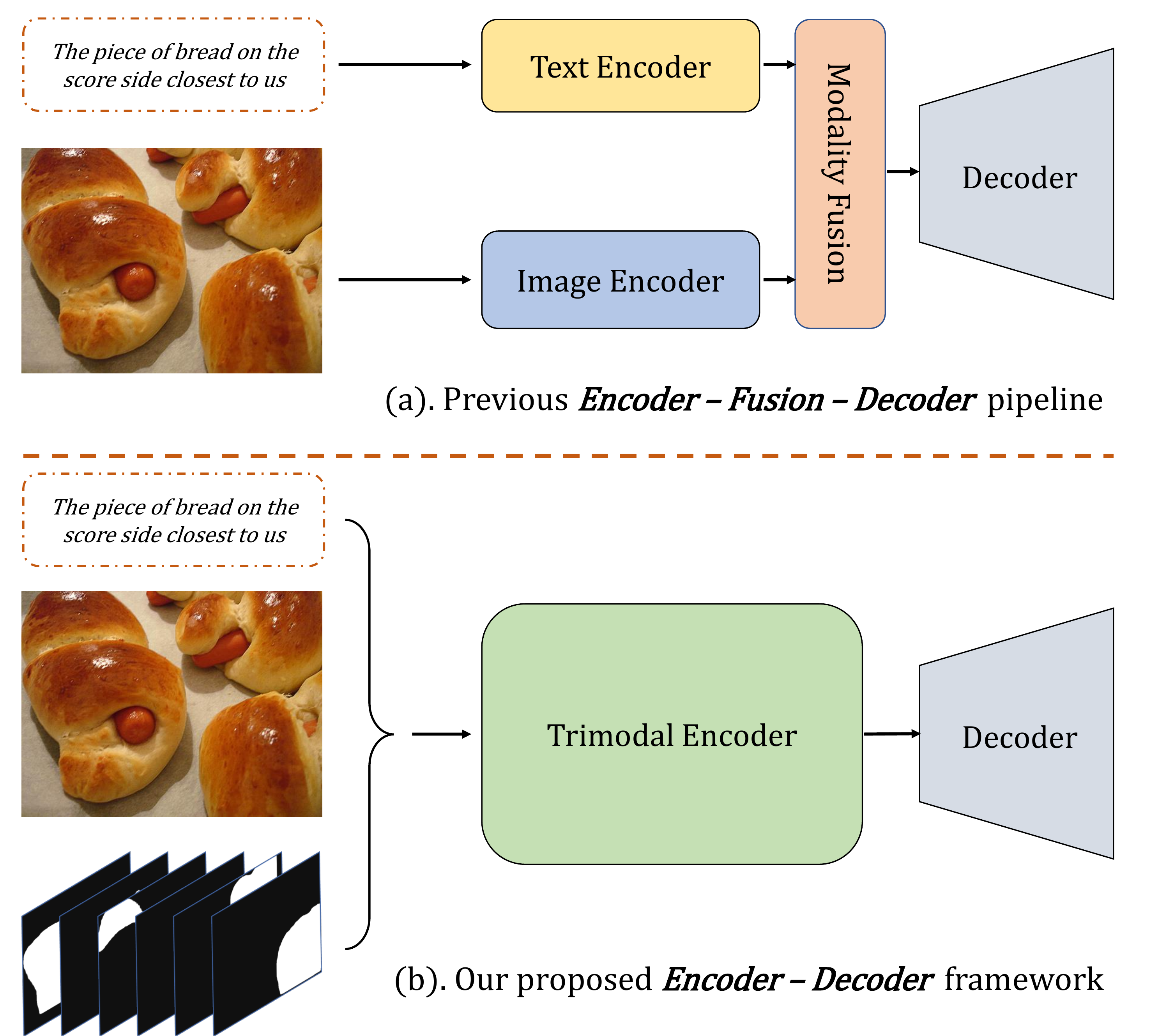}
   \caption{Comparison between (a) traditional \textit{encoder-fusion-decoder} pipeline and (b) our proposed \textit{encoder-decoder} pipeline which considers referring segmentation as a deep modality interaction task and introduces mask modality.}
   \label{fig:comparison}
\end{figure}

The \textit{de facto} paradigm to tackle the referring segmentation task is the \textit{encoder-fusion-decoder} pipeline, which first separately extracts image features and text features from two independent uni-modal encoders respectively, then fuses these representations together in a joint embedding space, finally decodes the target mask~(Fig.~\ref{fig:comparison}(a)). Prior arts following this pipeline have achieved substantial success but still suffer from three limitations: 
(1) Insufficient modality interaction, especially on low-level features: the modality fusion module is always applied to high-level representations that are learned independently from each modality. A series of works~\cite{ye2019cross, chen2019see, hu2020bi, hui2020linguistic, huang2020referring} have been proposed to focus on how to effectively fuse high-level features together after separate extraction, but these works may suffer from low-level detail information loss. Since referring segmentation is a pixel-level task, we believe that the low-level features like texture, colors and shapes in images and the features of the words in the language are important for precisely recognizing the target object, while only fusing the high-level features of deep uni-modality encoders always loses the low-level interaction. Therefore, fusing information from different modalities at both early and later stages is necessary to promote cross-modal representation for this task.
(2) Unaligned pre-trained tasks: the uni-modal encoders are often pre-trained on uni-modal tasks different from the target task. In detail, the visual feature extractors are usually pre-trained with image classification task on ImageNet~\cite{deng2009imagenet} and the language encoders are usually randomly initialized or pre-trained on language classification and generation tasks~\cite{devlin2018bert}. The pre-trained tasks guarantee network's ability on feature extraction in specific modality but inevitably bring inconsistency across different modalities for the cross-modal representations learning in referring image segmentation.
(3) Incomplete utilization of instance-level features: visual embeddings are always treated equally in terms of every location without highlighting in instances. Most of the previous methods~\cite{chen2019see, luo2020cascade, luo2020multi, ye2019cross} in this area directly use the global image representations without considering the instance-level features. However, for referring segmentation, the instance-level features should be highlighted since the referent in expression is often prone to describe instances. 
Some works~\cite{jing2021locate, yu2018mattnet} focus on the instance-level features by using regions or heatmaps, which are in a coarse and implicit manner.

To address the above limitations, this paper proposes \netname, a unified trimodal \textbf{Ma}sk-\textbf{I}mage-\textbf{L}anguage framework for joint image, language and mask learning, as illustrated in Fig.~\ref{fig:comparison}(b). 
\netname\ is mainly different from previous methods in two aspects. First, \netname\ models the referring segmentation task as a deep modality interaction problem with a more concise and effective \textit{encoder-decoder} pipeline. We employ a unified transformer structure as the encoder to fuse features from the input of each modality directly. This pipeline could directly handle the first two limitations since thorough cross-modal interaction is performed in the encoder and no uni-modal pre-training is needed anymore by discarding the deep single-modal encoders. We could adopt multi-modal pre-trained weights for our model, providing better aligned representations for this task compared to pre-training on unimodal tasks. 
The second different aspect is that our \netname\ is a trimodal framework rather than previous bimodal structure by explicitly considering the instance-level object masks as a modality. Intuitively, explicitly introducing the pre-segmented mask information could help the images and languages to pay more attentions to the instance-level features through modality interaction and strengthen the image features for finer prediction. To investigate this, we propose to regard the masks as a kind of modality, then we could build deep interactions among masks, language and image features at different feature levels. 
Moreover, in the decoder, we adaptively process all the candidate mask features to select the most relevant and informative one and compensate the image features with it as a visual prior to obtain the final prediction.

Our contributions are three folds: 
\begin{itemize}
    \vspace{-0.5em}
    \item We propose \netname, which is a new unified \textbf{Ma}sk-\textbf{I}mage-\textbf{L}anguage trimodal framework for referring image segmentation. The proposed deep modality interaction structure greatly simplifies the whole pipeline and sufficiently strengthens the interaction of multi-modal features.
    \vspace{-0.2em}
    \item For the first time, we propose to introduce object masks as an additional modality for this task. Together with the adaptive mask selection strategy, we make full use of the mask information and find that it indeed benefits a lot to this task.
    \vspace{-0.2em}
    \item The proposed \netname\ set a new state-of-the-art on all datasets with significant gains comparing to previous systems in the literature. Specifically, \netname\ outperforms the prior leading methods by 3\%-10\% on RefCOCO, RefCOCO+, and G-Ref. 
    \vspace{-0.5em}
\end{itemize}

%% file: contents/3-related_work.tex
\section{Related Work}
\label{related}
\subsection{Referring Image Segmentation}

Referring image segmentation aims at generating binary mask for referent described in given language expressions. Hu \textit{et al.}~\cite{hu2016segmentation} first proposes this task, and they construct an \textit{encoder-fusion-decoder} pipeline to tackle this problem. They first encode language and image features from two separate encoders, then fuse them by simple concatenation and leverage a fully convolutional network for segmentation. Following this paradigm, proceeding works mainly focus on how to effectively fuse extracted language and image features.
Recurrent network in \cite{liu2017recurrent, margffoy2018dynamic} and pyramid feature map in \cite{li2018referring} are utilized to excavate more semantic context for fusion. Luo \textit{et al.}~\cite{luo2020multi} proposes to learn referring segmentation and comprehension in a unified manner for better aligned representation.
Inspired by the prevalence of attention mechanism in computer vision field, researchers resort to attention mechanism for an effective fusion of multi-modal representations. Ye \textit{et al.}~\cite{ye2019cross} develops self-attention and cross-attention modules to capture long-range correlations between two modalities. Hu \textit{et al.}~\cite{hu2020bi} constructs two kinds of attention: language-guide visual attention and visual-guided language attention for two modalities separately. Recent \cite{jain2021comprehensive, ding2021vision} further employ transformer architecture in fusion stage for its better ability on feature interaction. Additionally, \cite{huang2020referring, hui2020linguistic, yang2021bottom} regard the fusion stage as a reasoning process. \cite{huang2020referring, hui2020linguistic} utilize graph structure to highlight correct entity and suppress other irrelevant ones. 
The aforementioned methods all follow the \textit{encoder-fusion-decoder} pipeline and focus on how to better fuse the high-level representations of different modalities. Feng \textit{et al.}~\cite{feng2021encoder} manage to fuse language features into visual encoder for more thorough feature interaction, but their fusion only takes place in last several layers. 

Unlike existing \textit{encoder-fusion-decoder} methods, our \netname~follows a more compact \textit{encoder-decoder} pipeline. With a unified transformer-based encoder, \netname\ encourages sufficient feature interaction to begin at the early stage of network, thus better aligning the representations across modalities. 

\textbf{Top-down perspective} is emphasized in another line of works~\cite{yu2018mattnet, chen2019referring, jing2021locate, liang2021clawcranenet} in referring image segmentation. They apply different methods to let network focus on instance-level features and model their relationships with language features, instead of conducting feature interaction with the whole image's feature map equally. Yu \textit{et al.}~\cite{yu2018mattnet} exploits modular relation scores into Mask R-CNN~\cite{he2017mask} framework for coarse region features, and Chen \textit{et al.}~\cite{chen2019referring} introduces cycle-consistency to further improve the performance. In \cite{jing2021locate} the top-down perspective is adopted implicitly by generating heat map as visual prior and concatenating it with image feature map. These methods utilize instance-level features in an implicit and coarse way. Recent ClawCraneNet~\cite{liang2021clawcranenet} in action segmentation field emphasizes top-down perspective as an object-level sensation that can capture high-level relations. However, their performance can be easily affected by the quality of the pre-trained segmentation model since they lack appropriate way to refine the final prediction.

 Different from them, we explicitly introduce object masks as an additional modality, which provides a strong prior and explicitly enhances the instance-level features for refined, precise results.

\subsection{Multi-modal learning}
Multi-modal has been an active research area in recent years. A comprehensive survey of multi-modal learning is beyond the scope of this paper, so we only briefly review several related works here, especially on recent transformer-based image-language representation learning.

Existing methods for image-language learning could be classified into two categories. The first line takes a dual-stream structure, which focuses on learning strong separate encoders for images and languages. Most of works~\cite{radford2021learning,jia2021scaling,li2019visual} in this type focus on learning strong representations with massive noisy web data. Methods~\cite{kim2021vilt,li2021align,wang2021simvlm} in the second categories use a sing-stream framework. They mainly research on modeling the modality interaction after the uni-modal encoders. We are inspired by a work~\cite{kim2021vilt} in this line, which designs its entire network as a transformer for efficiency. In this paper, we regard referring image segmentation as a deep multi-modal interaction task and build our network upon existing visual-language pre-training models instead of separate pre-trained image and language encoders. We further introduce the global instance masks as another modality to enhance the instance features, which dramatically improves the performance.

%% file: contents/4-methods_new.tex
\section{Method}
\label{method}

\begin{figure*}[h]
  \centering
  \includegraphics[width=0.95\linewidth]{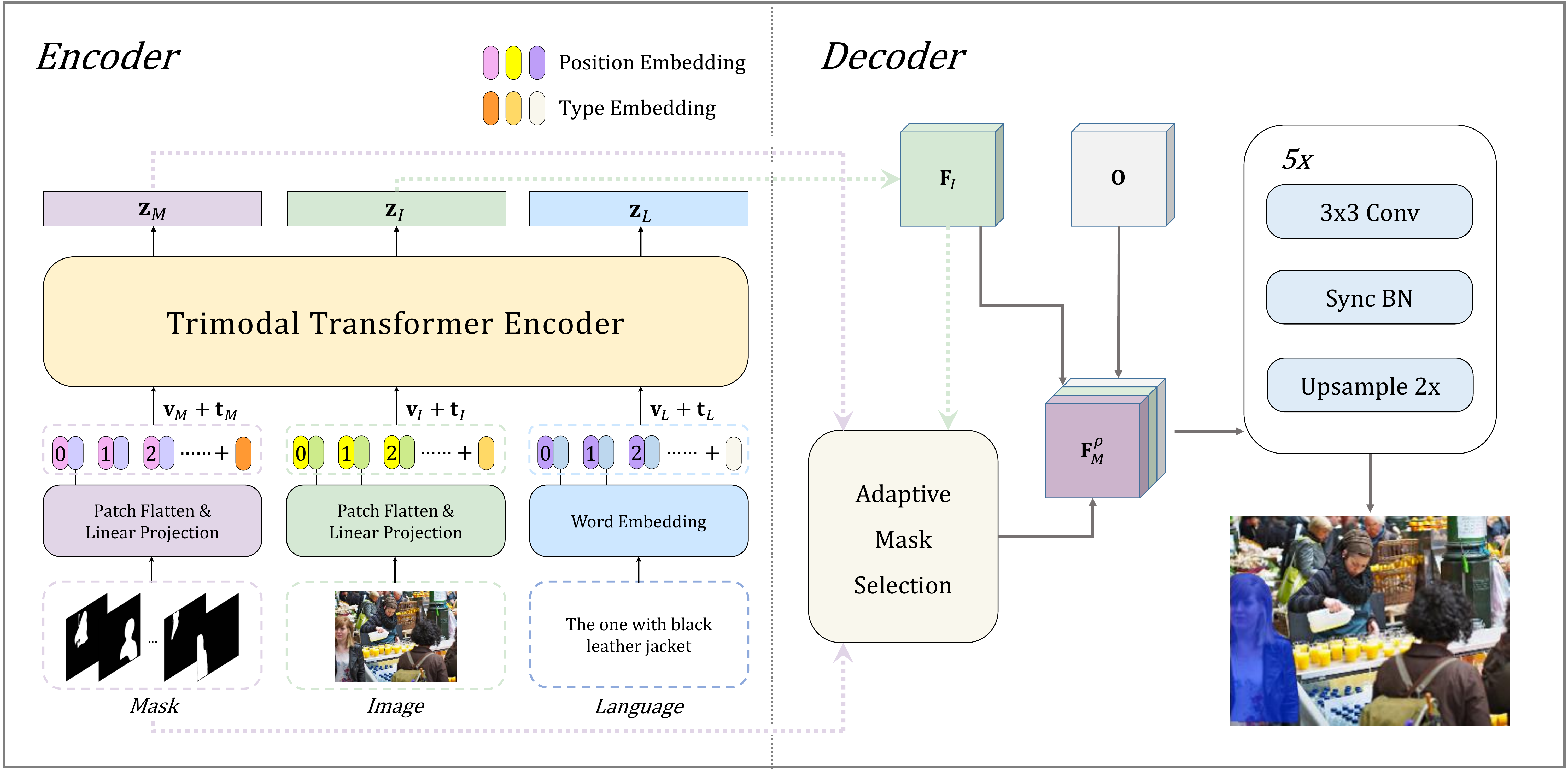}
  \caption{Overview of the proposed \netname. The trimodal encoder fuses feature from the input of each modality directly, jointly learning cross-modal representations among the mask, image and language modality. The decoder produces the segmentation results based on interacted representations with an adaptive mask selection strategy and a segmentation head.}
  \label{fig:overview}
\end{figure*}

The overall architecture of \netname\ is shown in Fig.~\ref{fig:overview}. It first takes the image, the referring expression and the corresponding instance masks as input, then feeds them into the proposed trimodal transformer encoder for sufficient multi-modality interaction, and finally generates the mask prediction by the decoder. In the following,  we will first briefly revisit the previous \textit{encoder-fusion-decoder} paradigm and present our compact \textit{encoder-decoder} pipeline (Sec.~\ref{methods:encoder-decoder}). Then we give an in-depth illustration of our core trimodal transformer encoder (Sec.~\ref{methods:trimodal encoder}). In the end, we will illustrate the decoder with the proposed mask selection strategy in detail (Sec.~\ref{methods:decoder}).

\subsection{Encoder-Decoder Pipeline}
\label{methods:encoder-decoder}

For referring image segmentation task, the input typically consists of an image $\mathbf{I} \in \mathbb{R}^{H \times W \times 3}$ and its corresponding language expression $\mathbf{L} = \left\{l_t\right\}_{t=1,\dots, T}$, where $H$ and $W$ denote the height and width of the input image. $T$ indicates the length of the expression. Previous methods always use an \textit{encoder-fusion-decoder} paradigm, which first adopts two uni-modal encoders~(\normalem{\eg}~ResNet~\cite{he2016deep} and GRU~\cite{chung2014empirical}) to extract images features $\mathbf{E}_I$ and languages features $\mathbf{E}_L$ separately, and then designs a modality fusion module to fuse representations from different modalities to obtain the fused features $\mathbf{F}$. In the end, $\mathbf{F}$ is fed into a decoder to generate the final segmentation prediction $\mathbf{P}$. This paradigm can be formulated as three steps:

\begin{equation}
\begin{split}
    & \mathbf{E}_I=\texttt{Encoder}_I(\mathbf{I}), \mathbf{E}_L =  \texttt{Encoder}_L(\mathbf{L}); \\
    & \mathbf{F} = \texttt{Fusion}(\mathbf{E}_I, \mathbf{E}_L); \\
    & \mathbf{P} = \texttt{Decoder}(\mathbf{F}) \\
\end{split}
\end{equation}

The above paradigm is faced with several limitations like the lack of interactions in low-level uni-modal features and the misalignment in the pre-trained uni-modal tasks and the target multi-modal task. To resolve these problems, we propose \netname\ to combine the original uni-modal encoders and the modality fusion module into one unified transformer encoder, resulting in a more succinct \textit{encoder-decoder} paradigm. It encourages the cross-modal feature interaction to start at a very early stage of the network. This is more effective for the low-level feature fusion, especially for this pixel-level segmentation task. Formally, the \textit{encoder-decoder} pipeline has two steps:

\begin{equation}
\begin{split}
    & \mathbf{F} = \texttt{Encoder}(\mathbf{I}, \mathbf{L}); \\
    & \mathbf{P} = \texttt{Decoder}(\mathbf{F}) \\
\end{split}
\label{equation:pipline}
\end{equation}

\subsection{Trimodal Transformer Encoder}
\label{methods:trimodal encoder}
Instance-level features are essential for the referring segmentation task since the referent in expression is often prone to describe specific instances~\cite{jing2021locate, yu2018mattnet,liang2021clawcranenet}. However, there are only a few methods~\cite{jing2021locate, yu2018mattnet} in this area focusing on this point and their usages are usually coarse and implicit, such as using regions or heatmaps. This paper proposes to highlight the instance-level features by explicitly leveraging much more delicate binary masks $\mathbf{M}$=$\left\{m_k\right\}_{k=1,\dots, K}$ ($K$ instances in an image), which could be directly generated by an off-the-shelf instance segmentation model~\cite{he2017mask}. 

Specifically, \netname\ introduces the generated masks as a modality and makes the encoder a trimodal one for joint learning with mask, image and language modalities. Then, the first step in Eq.~\ref{equation:pipline} becomes:
\begin{equation}
\mathbf{F}^* = \texttt{Encoder}^*(\mathbf{M}, \mathbf{I}, \mathbf{L})
\end{equation}
We adopt the vision transformer structure proposed in ViT~\cite{dosovitskiy2020image} as our encoder to handle the trimodal inputs. As shown in Fig.~\ref{fig:overview}, the input image $\mathbf{I} \in \mathbb{R}^{H \times W \times 3}$ is first sliced into patches and flattened as $\mathbf{\hat{I}} \in \mathbb{R}^{N\times (3P^2)}$, where $P$ is the patch size and $N = HW/P^2$. Then, followed by the linear projection and added with the positional embedding, the input image is embedded into $\mathbf{v}_I \in \mathbb{R}^{N\times d}$, where $d$ is the feature dimension of the transformer. A similar procedure is applied on the corresponding $K$ masks. They are then embedded into $\mathbf{v}_M\in \mathbb{R}^{N' \times d}$, where $N'=\sum_{k=1}^K H_k W_k / P^2$ and $\{H_k, W_k\}$ denotes the height and width of valid area in $k$-th mask.
The referring language is embedded to $\mathbf{v}_L\in \mathbb{R}^{T\times d}$ with a word embedding matrix and its position embedding matrix. After that, the embeddings of the three modalities are summed with their corresponding modality type embeddings respectively and then concatenated into a combined representation $\mathbf{z}^0\in \mathbb{R}^{(N' + N + T)\times d}$ before sending to the transformer:
\begin{equation}
    \mathbf{z}^0 = \left[\mathbf{v}_M + \mathbf{t}_{M};\mathbf{v}_I + \mathbf{t}_{I};\mathbf{v}_L + \mathbf{t}_{L}\right]
\end{equation}
where $\mathbf{t}_{M}, \mathbf{t}_{I}$ and $\mathbf{t}_{L}$ are the type embeddings of the three modalities respectively.

The trimodal transformer encoder consists of $D$ stacked blocks and each block contains a multi-head self-attention (MSA) layer and a multi-layer perceptron (MLP) layer, so the output of the $d$-th block is:

\begin{equation}
\begin{split}
    & \mathbf{\hat{z}}^d = \texttt{MSA}(\texttt{LN}(\mathbf{z}^{d-1})) + \mathbf{z}^{d-1}, \\
    & \mathbf{z}^d = \texttt{MLP}(\texttt{LN}(\mathbf{\hat{z}}^d)) + \mathbf{\hat{z}}^d. \\
\end{split}
\end{equation}
Here LN denotes layer normalization~\cite{ba2016layer}. The contextualized vector sequence $\mathbf{z}^d$ is iteratively updated through blocks, thus facilitating sufficient feature interaction across modalities due to the self-attention mechanism. The final output feature vectors of the trimodal encoder, $\mathbf{z}^D$, can be divided back to features of different modalities as $\left\{\mathbf{z}_M, \mathbf{z}_I, \mathbf{z}_L\right\}$. 

Now the features of each modality are fully attended with the information from others via the trimodal transformer encoder.
Finally, we reshape the image and mask features back to the original 2-D image size and obtain the encoder's output $\mathbf{F}^*$=$\{\mathbf{F}_I,\mathbf{F}_M\}$. Note that the dimensions are $\mathbf{F}_I$$\in$$\mathbb{R}^{H' \times W' \times d}$ and $\mathbf{F}_M$=$\left\{\mathbf{F}_M^k\in \mathbb{R}^{H' \times W' \times d} \right\}_{k=1,\cdots,K}$, where $H'$=$H/P$ and $W'$=$W/P$ due to the patch division.

\subsection{Decoder}
\label{methods:decoder}
After sufficient feature interaction in the encoder, the output image and mask features can be regarded as two types of representations containing rich information across different modalities. The image features include the whole image's semantic content and the instance-level mask features focus on each instance's representation. In order to generate precise segmentation results, our decoder needs to combine features from both two aspects. Specifically, we propose an adaptive selection strategy to process various numbers of mask features and choose the most informative one while discarding others. A segmentation head is placed afterward to produce results based on the combination of the whole image features and selected instance-level features. 

\begin{figure}[t!]
  \centering
  \includegraphics[width=\linewidth]{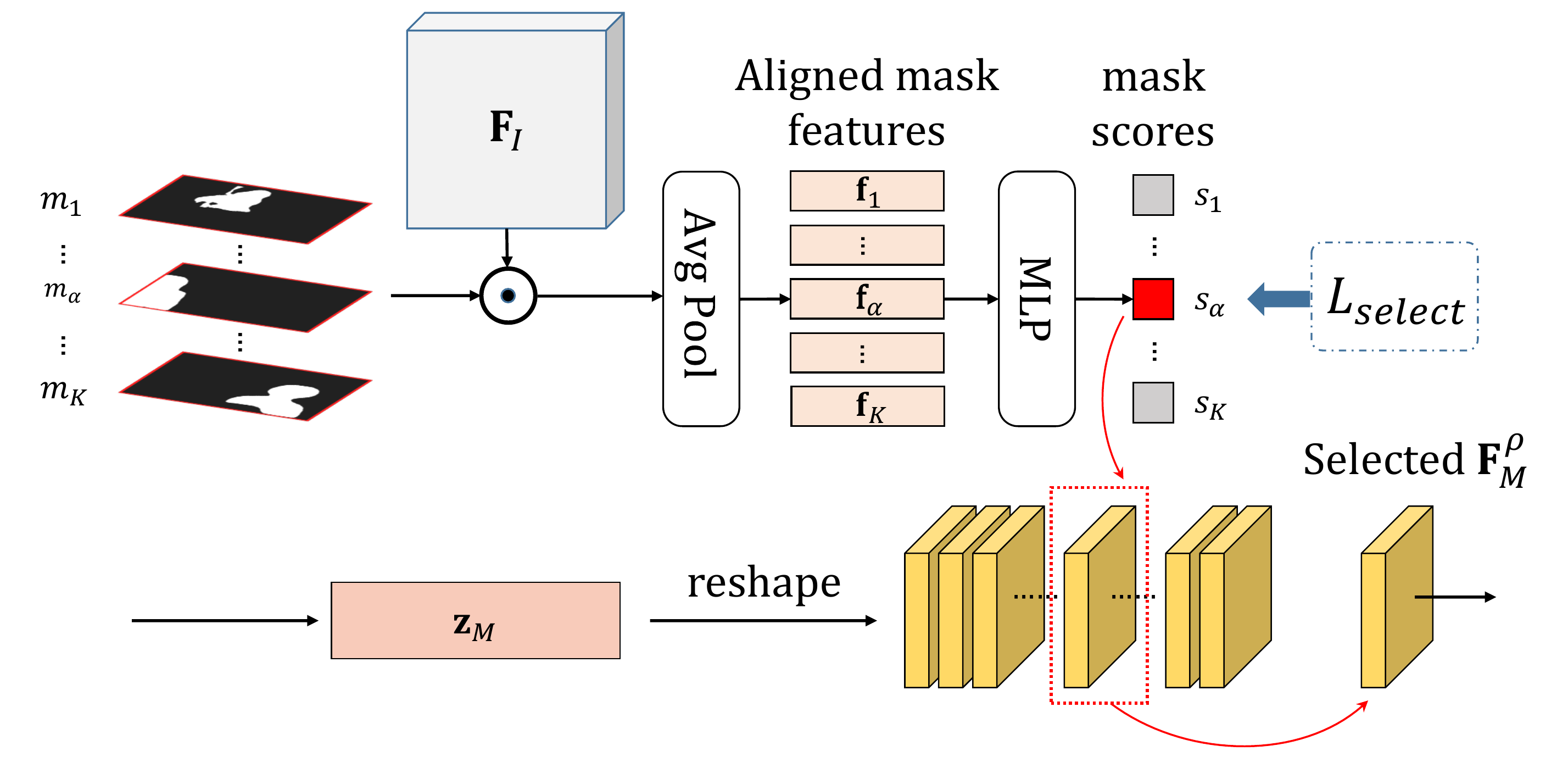}
  \caption{Proposed Adaptive Mask Selection Strategy.}
  \label{fig:module}
\end{figure}

\textbf{Adaptive Mask Selection Strategy.} 
It is evident that using the information of all masks is redundant and utilizing all of them will bring extra distractors, as demonstrated in the experiment (Sec.~\ref{section:ab_exp}). Therefore, we design an adaptive mask selection strategy to figure out the most informative mask features. As illustrated in Fig.~\ref{fig:module}, to make a fair comparison and choice in a common shared embedding space, our \netname\ first crops features in the image feature maps (we use the word ``aligned" for this type of feature for clarification) corresponding to the mask areas. In detail, for the $k$-th mask $m_k$, we first obtain the aligned feature vector $\mathbf{f}_k$ by average pooling on the features of the mask-cropped areas, and further use an MLP layer to derive a corresponding score $s_k$ for every mask. Mathematically, this process could be achieved by,

\begin{equation}
    \mathbf{f}_k = \texttt{AVG}(\mathbf{F}_I \odot m_k),\quad s_k = \texttt{MLP}(\mathbf{f}_k)
\end{equation}
Then, we could select the most informative mask feature $\mathbf{F}_M^{\rho } \in \mathbb{R}^{H' \times W' \times d}$ according to the largest score for further processing, where $\rho$ is the index of the maximum score.

Specifically, we consider a mask feature as the most informative one if it is the most similar to the ground truth. To let the network learn how to distinguish the most relevant instance mask among others, we set a training objective to maximize the predicted score for the mask which has the largest overlap with the ground truth mask wile minimizing the other scores. In detail, we first compute the intersection-over-union~(IoU) between the ground truth mask and each input candidate mask. Then, a contrastive loss is applied to the generated scores:

\begin{equation}
    L_{select} = -\log{\frac{\exp \left(s_{\alpha}\right)}{\Sigma_{k=1}^K \exp\left(s_k\right)}}.
    \label{eq:select_loss}
\end{equation}
where $\alpha$ is the index of the mask which has the largest IoU with the ground truth.

Next, the selected mask is used to supply the global image feature maps $\mathbf{F}_I$. 
More specifically, we first use two 1$\times$1 convolution layers to reduce the channels by a factor of $r$ to ease the computing cost on image feature map $\mathbf{F}_I$ and $\mathbf{F}_M^{\rho}$ respectively. Then we concatenate them with a $2$-D spatial coordinate map $\mathbf{O}\in \mathbb{R}^{H'\times W'\times 2}$ and obtain the concatenated feature maps $\hat{\mathbf{F}}$,

\begin{equation}
    \hat{\mathbf{F}} = \texttt{Concat}(\mathbf{F}_I, \mathbf{F}_M^{\rho}, \mathbf{O})
\end{equation}
$ \hat{\mathbf{F}}$ will be sent to the next step, the segmentation head.

\textbf{Segmentation Head.} To perform pixel-level segmentation in original space, we adopt a progressive up-sampling segmentation structure~\cite{zheng2021rethinking}. It contains multiple stacked blocks, and the architecture of each block is: $3\times$$3$ conv + sync batch norm + ReLU + $2\times$ bi-linear up-sampling, and one $1\times$$1$ conv attached at last block as shown in Fig.~\ref{fig:overview}. We use $\text{log}_{2}P$ blocks to recover the input resolution. After that, the decoder generates the final mask prediction $\mathbf{P} \in \mathbb{R}^{H\times W \times 1}$, which has the identical shape to the original image.
We use the Focal loss~\cite{lin2017focal} and DICE loss~\cite{milletari2016v} to supervise the generated mask, and sum up with $\lambda L_{select}$ (Eq.~\ref{eq:select_loss}) as the final training objective of our proposed \netname, where $\lambda$ is used to balance the losses.

%% file: contents/5-experiments.tex
\section{Experiments}
\label{exp}

\begin{table*}[t!]
   \centering
  \small
   \setlength{\tabcolsep}{2.8mm}{\begin{tabular}{@{}lccccccccc@{}}
      \toprule
      \multirow{2}[4]{*}{} & \multicolumn{3}{c}{RefCOCO} & \multicolumn{3}{c}{RefCOCO+} & \multicolumn{3}{c}{G-Ref} \\
      \cmidrule{2-10}
                                     & val   & test A & test B & val   & test A & test B & val (U)   & test (U)  & val(G)   \\
      \midrule
      RMI~\cite{liu2017recurrent}    & 45.18 & 45.69 & 45.57 & 29.86 & 30.48 & 29.50 & -     & -     & 34.52 \\
      DMN~\cite{margffoy2018dynamic} & 49.78 & 54.83 & 45.13 & 38.88 & 44.22 & 32.29 & -     & -     & 36.76 \\
      ASGN~\cite{qiu2019referring}   & 50.46 & 51.20 & 49.27 & 38.41 & 39.79 & 35.97 & -     & -     & 41.36 \\
      RRN~\cite{li2018referring}     & 55.33 & 57.26 & 53.93 & 39.75 & 42.15 & 36.11 & -     & -     & 36.45 \\
      MAttNet~\cite{yu2018mattnet}   & 56.51 & 62.37 & 51.70 & 46.67 & 52.39 & 40.08 & 47.64 & 48.61 & -     \\
      CMSA~\cite{ye2019cross}        & 58.32 & 60.61 & 55.09 & 43.76 & 47.60 & 37.89 & -     & -     & 39.98 \\
      CAC~\cite{chen2019referring}   & 58.90 & 61.77 & 53.81 & -     & -     & -     & 46.37 & 46.95 & 44.32 \\
      STEP~\cite{chen2019see}        & 60.04 & 63.46 & 57.97 & 48.19 & 52.33 & 40.41 & -     & -     & 46.40 \\
      BRINet~\cite{hu2020bi}         & 60.98 & 62.99 & 59.21 & 48.17 & 52.32 & 42.11 & -     & -     & 48.04 \\
      CMPC~\cite{huang2020referring} & 61.36 & 64.53 & 59.64 & 49.56 & 53.44 & 43.23 & -     & -     & 39.98 \\
      LSCM~\cite{hui2020linguistic}  & 61.47 & 64.99 & 59.55 & 49.34 & 53.12 & 43.50 & -     & -     & 48.05 \\
      MCN~\cite{luo2020multi}        & 62.44 & 64.20 & 59.71 & 50.62 & 54.99 & 44.69 & 49.22 & 49.40 & -     \\
      BUSNet~\cite{yang2021bottom}   & 62.56 & 65.61 & 60.38 & 50.98 & 56.14 & 43.51 & -     & -     & 49.98 \\
      EFN~\cite{feng2021encoder}     & 62.76 & 65.69 & 59.67 & 51.50 & 55.24 & 43.01 & -     & -     & \underline{51.93} \\
      CGAN~\cite{luo2020cascade}     & 64.86 & 68.04 & 62.07 & 51.03 & 55.51 & 44.06 & 51.01 & 51.69 & 46.54 \\
      JRNet~\cite{jain2021comprehensive} & 65.32 & \underline{68.56} & 62.04 & 52.75 & 58.46 & 44.12 & - & - & 48.95 \\
      LTS~\cite{jing2021locate}      & 65.43 & 67.76 & \underline{63.08} & 54.21 & 58.32 & 48.02 & \underline{54.40} & 54.25 & - \\
      VLT~\cite{ding2021vision}      & \underline{65.65} & 68.29 & 62.73 & \underline{55.50} & \underline{59.20} & \underline{49.36} & 52.99 & \underline{56.65} & 49.76 \\
      \midrule
      \textbf{\netname}~(416$\times$416)               & 69.38 & 71.31 & 66.76 & 61.02 & 64.68 & 54.86 & 60.86 & 61.39 & 60.26 \\
      \textbf{\netname}      & \textbf{70.13} & \textbf{71.71} & \textbf{66.92} & \textbf{62.23} & \textbf{65.92} & \textbf{56.06} & \textbf{62.45} & \textbf{62.87} & \textbf{61.81} \\
      \bottomrule
   \end{tabular}}%
   \caption{Comparison of state-of-the-art methods with ours on RefCOCO, RefCOCO+ and G-Ref with the IoU metric. Previous best results are underlined for emphasize. U: UMD split, G: Google split.}
   \label{tab:results}%
\end{table*}%

We evaluate our proposed method on three commonly used benchmark datasets for referring image segmentation, including RefCOCO~\cite{yu2016modeling}, RefCOCO+~\cite{yu2016modeling} and G-Ref~\cite{mao2016generation, nagaraja2016modeling}, with the mask intersection-over-union~(IoU) and Precision@$X$ as evaluation metrics. The details of the datasets and evaluation metrics are shown in the appendix.

\subsection{Implementation Details}

Unlike previous methods which use pre-trained deep unimodal encoders for separate feature extraction, here we first adopt a \texttt{bert-base-uncased} tokenizer to tokenize the text inputs and a publicly available pre-trained modality transformer~\cite{kim2021vilt} for joint image, expression and masks feature extraction and modality interaction as mentioned in Sec.~\ref{methods:trimodal encoder}. The trimodal transformer adopts the original ViT~\cite{dosovitskiy2020image} structure, the patch size $P=32$, the embedding dimension $d$=768, with $D$=12 blocks and each with $12$ attention heads, all following the original setting. The decoder has $5$ stacked blocks to resize the intermediate feature map size to $32\times$. We adopt two settings for image resolution: one is to resize all images to $416\times416$ as in previous works~\cite{luo2020multi, yu2018mattnet, ding2021vision}, the other is to first resize the shorter edge to $384$ and limit the longer edge to under $640$ while preserving the aspect ratio as~\cite{kim2021vilt}. $r$ in the decoder is set to 3 and $\lambda$ is set to 0.1 for loss balance. Our off-the-shelf segmentation model is a ResNet-50 Mask R-CNN~\cite{he2017mask} pre-trained on instances in train splits of RefCOCO/RefCOCO+/G-Ref. And the maximum length for expression is set to 15 for RefCOCO and RefCOCO+ and 20 for G-Ref. 

We use AdamW~\cite{loshchilov2017decoupled} optimizer with a base learning rate of $10^{-4}$ and weight decay of $10^{-2}$. The training phase contains 10 epochs and the learning rate is warmed up for $10\%$ of the total training steps and decayed linearly to zero for the rest of training. We train our network on 4 NVIDIA RTX3090 GPUs with a batch size of 128.

\begin{table*}[t!]
    \centering
    \small
    \setlength{\tabcolsep}{2mm}
    \begin{tabular}{@{}ccccccccc@{}}
    \toprule
       Variants & Encoder Modality & Decoder Modality & Pr@$0.5$ & Pr@$0.6$ & Pr@$0.7$ & Pr@$0.8$ & Pr@$0.9$ & IoU \\
    \midrule
      1& Image + Language & Image & 69.10 & 62.52 & 53.08 & 35.31 & 8.75 & 59.20 \\
    \midrule
      2& Mask + Language & Mask & 55.62 & 51.52 & 42.45 & 24.48 & 04.27 & 46.19 \\
    \midrule
      3& Image + Mask + Language & Image & 69.24 & 63.23 & 54.26 & 37.27 & 10.92 & 59.52 \\
       4&Image + Mask + Language & Mask & 67.22 & 63.51 & 57.08 & 39.28 & 9.63 & 56.51 \\
      5& Image + Mask + Language & Image + Mask & \textbf{72.35} & \textbf{67.79} & \textbf{60.54} & \textbf{45.82} & \textbf{15.29} & \textbf{62.23} \\
     \midrule
      6& Image + Mask + Language~\dag & Image + Mask & 61.89 & 58.35 & 52.47 & 40.52 & 13.66 & 53.96 \\
    \bottomrule
    \end{tabular}
    \caption{Ablation study of the pipeline and modality configuration on the validation set of RefCOCO+. Encoder Modality denotes input modalities of transformer encoder, while Decoder Modality denotes which modalities output of the encoder are fed into the decoder. \dag~: without multi-modal pre-train weight~\cite{kim2021vilt}.}
    \label{tab:ablation1}
\end{table*}

\begin{table}[t!]
    \centering
    \small
    \setlength{\tabcolsep}{0.5mm}
    \begin{tabular}{@{}ccccccc@{}}
    \toprule
         Methods
         & Pr@$0.5$ & Pr@$0.6$ & Pr@$0.7$ & Pr@$0.8$ & Pr@$0.9$ & IoU \\
    \midrule
       Mean & 70.91 & 65.34 & 58.02 & 42.95 & 13.66 & 61.26 \\
       Maximum & 71.21 & 66.22 & 58.54 & 43.58 & 13.47 & 61.30 \\
       Weighted Sum & 71.07 & 66.13 & 59.71 & 44.34 & 14.25 & 61.36 \\
    \midrule
       Adaptive Select & \textbf{72.35} & \textbf{67.79} & \textbf{60.54} & \textbf{45.82} & \textbf{15.29} & \textbf{62.23} \\
    \bottomrule
    \end{tabular}
    \caption{Ablation study of mask selection strategies.}
    \label{tab:ablation2}
\end{table}

\subsection{Comparison with State-of-the-art}
We compare the performance of our proposed \netname\ with state-of-the-art methods on three widely-used datasets in Tab.~\ref{tab:results}. Following prior works~\cite{ye2019cross,yang2021bottom,ding2021vision}, we report the IoU score here for page limitation and present the full results in the appendix. There are two input resolution configurations implemented as mentioned above. We show the results of both the two versions in Tab.~\ref{tab:results} and adopt the fixed aspect ratio in our final version.

 It can be seen that our proposed method consistently outperforms existing methods across all three benchmark datasets by large margins. In particular, \netname\ is higher than existing best results with obvious gains of about 3\%-5\% on the three sets of the RefCOCO. For the more difficult dataset RefCOCO+, \netname\ achieves significant improvements on all the splits compared with the previous top results by almost 7\%. Furthermore, for G-Ref with more extended and complex expressions, our method yields notable boosts of 8.05\%, 6.22\% and 9.88\% on three splits compared to previous best results. 
 
 Notably, \netname\ achieves greater performance gains than previous methods as the complexity of the dataset increases. This success stems from two aspects: (1) the deep modality interaction of \netname, where the long and complex sentences could sufficiently interact with the features of masks and images at different levels, then the most useful information for the target object could be easily found. (2) the introduction of the mask modality is beneficial for enhancing the informative instance-level features and build tight relationships between these instance features with the expressions to figure out the most possible candidate.

\begin{figure*}[t!]
  \centering
  \includegraphics[width=0.9\linewidth]{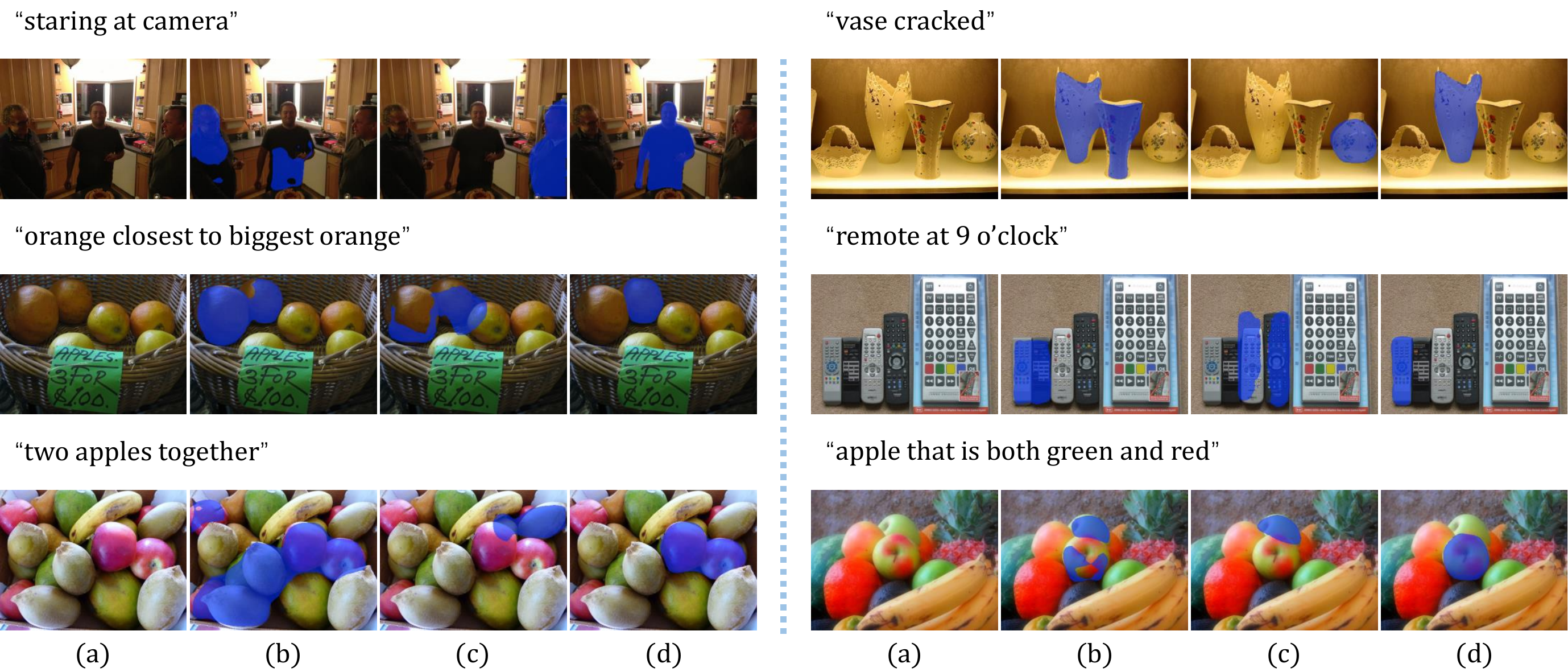}
  \vspace{-0.2cm} \\
  \caption{Examples of the segmentation results with different input modalities. Given referring expression and (a)~image, the other three columns compare the results from (b)~\netname\ w/o mask modality, (c)~\netname\ w/o image modality and (d)~\netname. Best view in color.}
  \label{fig:vis_comparison}
\end{figure*}

\begin{figure}[t!]
  \centering
  \includegraphics[width=\linewidth]{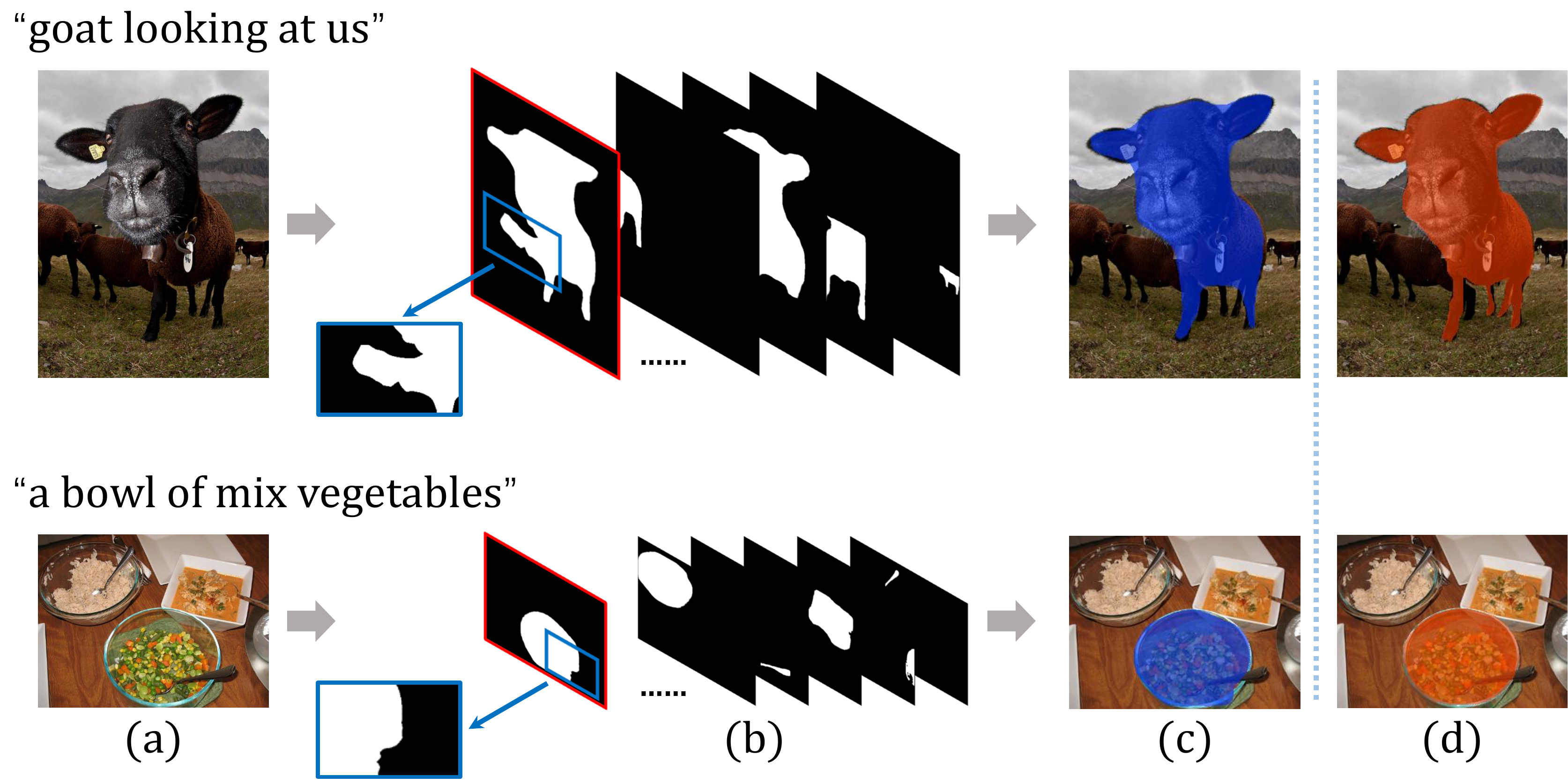}
  \vspace{-0.5cm} \\
  \caption{Visualization of the adaptive mask selection results. (a)~Input image and expression, (b)~input masks, the one with largest score highlighted, (c)~\netname\ output, (d)~ground truth.}
  \label{fig:vis_score}
\end{figure}

\subsection{Ablation Study}\label{section:ab_exp}

We conduct a series of ablation experiments to validate our method on the validation split of the RefCOCO+.

\textbf{Is the \textit{encoder-decoder} pipeline helpful?} To demonstrate the benefit of our pipeline, we implement variant 1 and variant 6 in Tab.~\ref{tab:ablation1}, where the former only uses the image and language modalities with a bimodal transformer encoder and the latter removes the multi-modal pre-trained weights from \netname. Variant 1 obtains 59.20\% of IoU score, which already surpasses all the methods in Tab.~\ref{tab:results} with a large margin. Variant 6 drops about 8\% compared to our \netname\ (but still comparable to the state-of-the-art). The results verify our original motivations that our \textit{encoder-decoder} pipeline could (1) empower the sufficient modality interaction and (2) take advantage of aligned multi-modal pre-training task. 

\textbf{Is the mask modality beneficial?} We demonstrate this problem with different configurations of the modalities of the encoder and the decoder, respectively. For the encoder, we construct three variants: variant 1~(image + language), variant 2~(mask + language) and variant 5~(image + mask + language). Note that for variant 2, we use the mask-cropped images to keep the original pixel information inside masks. The results between variant 1 and variant 5 demonstrate that introducing mask as an independent modality can bring $2.89\%$ performance gain~(IoU) since it enables the network pay more attention to the instance-level features. On the other hand, directly using mask modality like variant 2 leads to a large performance drop. We attribute it to the loss of the background features and the location information. We provide the visualization in Fig.~\ref{fig:vis_comparison} for better understanding. 
For the input of decoder, we investigate the influence of different modalities when given three modalities inputs in encoder part. Specifically, we have three variants: variant 3~(image), variant 4~(mask) and variant 5~(image + mask). From which we have 3 findings: (1) Simply sending the masks into encoder without any further process in decoder~(variant 3) brings little improvement against variant 1. (2) Only feeding mask representations to the decoder~(variant 4) causes the performance degradation since image features also matter and provide rich information. (3) Explicitly combining features from mask and image modality~(variant 5, i.e., our \netname) with the proposed select strategy achieves the best results. Therefore, the final conclusion is that the mask modality is beneficial when the modality for both encoder and decoder is properly configured. 

\textbf{Mask Selection Strategies.}
Tab.~\ref{tab:ablation2} compares different approaches for choosing the mask features from the numerous $K$ mask representations from the encoders. Four strategies are explored: Mean, Maximum, Weighted Sum and our proposed strategy. The first two are simple average pooling or max-pooling operations on the $K$ mask features. Both attempts could improve the performance compared with using no mask features as variant 3 in Tab.~\ref{tab:ablation1}. The Weighted Sum is the most related one to ours, where the difference is that it normalizes the learned scores then weighted sums all the mask features. This can bring minor improvement against the first two since it allows the network to learn where to focus and suppress influence from irrelevant areas. However, these strategies consider embeddings of all masks, thus the irrelevant areas' information can be misleading. Finally, our proposed adaptive selection strategy achieves the best results by directly using mask embedding with the highest score, which brings 2.57\% improvement against variant 3. The visualization of the chosen masks in Fig.~\ref{fig:vis_score} also proves the effectiveness of the proposed strategy.

\subsection{Visualization Results}
\label{vis}
In Fig.~\ref{fig:vis_comparison}, we show the qualitative comparison of our method and when lack of image modality or mask modality, \ie, variant 5 (\netname), variant 1 and 2 in Tab.~\ref{tab:ablation1}. Variant 1 mistakenly segments out referent and its surroundings, especially the distractors, because of the lack of instance-level mask modality. For instance, it fails to distinguish the target vase, fruits and remote with other adjacent objects belonging to the same categories. As for variant 2, which only has mask-cropped image areas as inputs and omits the background, its output can be obviously affected by the quality of the pre-segmented input masks and leading to completely wrong results like the examples in the visualization.

We also present the visualization of the mask selection results in Fig.~\ref{fig:vis_score}. The mask with the highest score is highlighted using the red frame. It can be seen that our proposed adaptive mask selection strategy can select the most relevant mask. Besides, when looking at the comparison between the \netname's final result and selected mask, it demonstrates the segmentation head with image feature map can amend flaws in the pre-generated mask. For instance, the selected mask in the first row of Fig.~\ref{fig:vis_score} segments a spare piece of an area on the left of the target goat, while the final output mask successfully removes this part.

%% file: contents/6-conclusion.tex
\section{Conclusion}
\label{conclusion}
In this paper, we present \netname, a unified network for referring image segmentation. There are two fundamental differences from existing works: (i) Unify the uni-modal feature extractors and fusion models as one deep modality interaction transformer, simplifying the pipeline from \textit{encoder-fusion-decoder} to \textit{encoder-decoder} and enabling sufficient modality interaction in multi-level features; (ii) Explicitly introduce the instance masks as an additional modality, constructing a trimodal framework rather than previous bimodal networks and intensifying the beneficial instance-level features. Experiments across diverse benchmarks of this task show that proposed \netname\ outperforms state-of-the-art methods with large margins (3\%-10\%), setting a new baseline for this task. 